\documentclass[runningheads]{llncs}
\usepackage[utf8]{inputenc}

\usepackage{color,xcolor}
\usepackage{epsfig}
\usepackage{graphicx}

\usepackage{adjustbox}
\usepackage{array}
\newcolumntype{?}[1]{!{\vrule width #1}}
\usepackage{booktabs}
\usepackage{colortbl}
\usepackage{float,wrapfig}
\usepackage{hhline}
\usepackage{multirow}
\usepackage{subcaption} 
\usepackage[labelfont={bf},labelsep={period},font={small}]{caption}

\let\llncssubparagraph\subparagraph
\let\subparagraph\paragraph
\usepackage[compact]{titlesec}
\let\subparagraph\llncssubparagraph

\usepackage{amsmath,amsfonts,amssymb}
\usepackage{bm}
\usepackage{nicefrac}
\usepackage{microtype}

\usepackage{changepage}
\usepackage{extramarks}
\usepackage{fancyhdr}
\usepackage{lastpage}
\usepackage{setspace}
\usepackage{soul}
\usepackage{xspace}

\usepackage{url}

\usepackage{algorithm, algorithmic}
\usepackage{enumerate}

\newcolumntype{L}[1]{>{\raggedright\let\newline\\\arraybackslash\hspace{0pt}}m{#1}}
\newcolumntype{C}[1]{>{\centering\let\newline\\\arraybackslash\hspace{0pt}}m{#1}}
\newcolumntype{R}[1]{>{\raggedleft\let\newline\\\arraybackslash\hspace{0pt}}m{#1}}

\newcommand{\sect}[1]{Section~\ref{#1}}

\newcommand{\eqn}[1]{Equation~\ref{#1}}
\newcommand{\fig}[1]{Figure~\ref{#1}}
\newcommand{\tbl}[1]{Table~\ref{#1}}

\newcommand{\ignore}[1]{}
\newcommand{\norm}[1]{\lVert#1\rVert}

\makeatletter
\DeclareRobustCommand\onedot{\futurelet\@let@token\@onedot}
\def\@onedot{\ifx\@let@token.\else.\null\fi\xspace}

\def\eg{\emph{e.g}\onedot} 
\def\ie{\emph{i.e}\onedot} 
 
 \def\vs{\emph{vs}\onedot}
 
\def\etal{\emph{et al}\onedot}
\makeatother

\definecolor{MyDarkBlue}{rgb}{0,0.08,1}
\definecolor{MyDarkGreen}{rgb}{0.02,0.6,0.02}
\definecolor{MyDarkRed}{rgb}{0.8,0.02,0.02}
\definecolor{MyDarkOrange}{rgb}{0.40,0.2,0.02}
\definecolor{MyPurple}{RGB}{111,0,255}
\definecolor{MyRed}{rgb}{1.0,0.0,0.0}
\definecolor{MyGold}{rgb}{0.75,0.6,0.12}
\definecolor{MyDarkgray}{rgb}{0.66, 0.66, 0.66}

\newcommand{\myparagraph}[1]{\vspace{1pt}\noindent{\bf #1.}}
\newcommand{\myackparagraph}[1]{\noindent{\bf #1:}}

\newcommand{\model}{PPD\xspace}

\begin{document}

\pagestyle{headings}
\mainmatter
\def\ECCV18SubNumber{1300}

\title{Physical Primitive Decomposition}

\author{Zhijian Liu\inst{1} \and
William T. Freeman\inst{1,2} \and
Joshua B. Tenenbaum\inst{1} \and
Jiajun Wu\inst{1}}
\institute{Massachusetts Institute of Technology \and
Google Research}

\maketitle

\begin{abstract}

Objects are made of parts, each with distinct geometry, physics, functionality, and affordances. Developing such a distributed, physical, interpretable representation of objects will facilitate intelligent agents to better explore and interact with the world. In this paper, we study \emph{physical primitive decomposition}---understanding an object through its components, each with physical and geometric attributes. As annotated data for object parts and physics are rare, we propose a novel formulation that learns physical primitives by explaining both an object's appearance and its behaviors in physical events. Our model performs well on block towers and tools in both synthetic and real scenarios; we also demonstrate that visual and physical observations often provide complementary signals. We further present ablation and behavioral studies to better understand our model and contrast it with human performance. 

\end{abstract}

\section{Introduction}
\label{sec:intro}

Humans use a hammer by holding its handle and striking its head, not vice versa. In this simple action, people demonstrate their understanding of \emph{functional parts}~\cite{Rivlin1995Recognition,Tenenbaum1994Functional}: a tool, or any object, can be decomposed into primitive-based components, each with distinct physics, functionality, and affordances~\cite{Gibson1977Affordances}. 

\begin{figure}[t]
\centering
\includegraphics[width=0.75\linewidth]{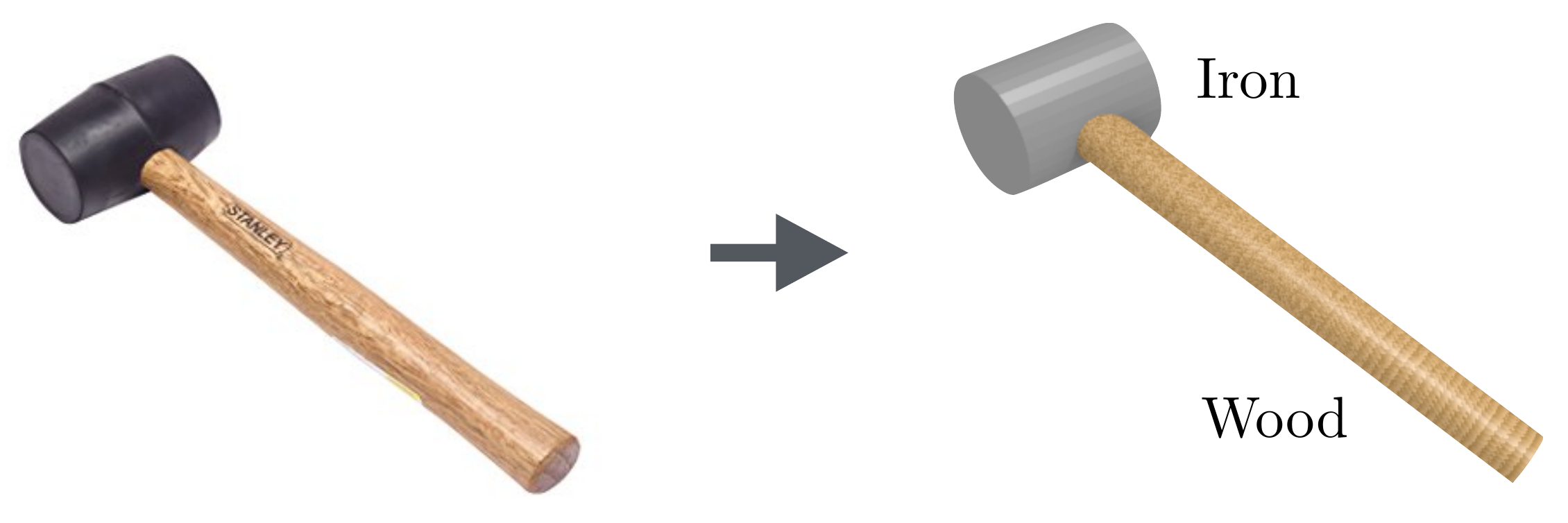}
\caption{A hammer (left) and its physical primitive decomposition (right).}
\label{fig:teaser}
\vspace{-6mm}
\end{figure}

How to build a machine of such competency? In this paper, we tackle the problem of \emph{physical primitive decomposition (\model)}---explaining the shape and the physics of an object with a few shape primitives with physical parameters. Given the hammer in \fig{fig:teaser}, our goal is to build a model that recovers its two major components: a tall, wooden cylinder for its handle, and a smaller, metal cylinder for its head.

For this task, we need a physical, part-based object shape representation that models both object geometry and physics. Ground-truth annotations for such representations are however challenging to obtain: large-scale shape repositories like ShapeNet~\cite{Chang2015Shapenet:} often have limited annotations on object parts, let alone physics. This is mostly due to two reasons. First, annotating object parts and physics is labor-intensive and requires strong domain expertise, neither of which can be offered by current crowdsourcing platforms. Second, there exist intrinsic ambiguity in the ground truth: it is impossible to precisely label underlying physical object properties like densities from only images or videos. 

Let's think more about what these representations are \emph{for}. We want our object  representation to faithfully encode its geometry; therefore, it should be able to explain our visual observation of the object's appearance. Further, as the representation models object physics, it should be effective in explaining the object's behaviors in various physical events.

Inspired by this, we propose a novel formulation that learns a part-based object representation from both visual observations and physical interactions. Starting with a single image and a voxelized shape, the model recovers the geometric primitives and infers their physical properties from texture. The physical representation inferred this way is of course rather uncertain; it therefore only serves as the model's prior of this physical shape. Observing object behaviors in physical events offers crucial additional information, as objects with different physical properties behave differently in physical events. This is used by the model in conjunction with the prior to produce its final prediction.

We evaluate our system for physical primitive decomposition in three scenarios. First, we generate a dataset of synthetic block towers, where each block has distinct geometry and physics. Our model is able to successfully reconstruct the physical primitives by making use of both appearance and motion cues. Second, we evaluate the system on a set of synthetic tools, demonstrating its applicability to daily-life shapes. Third, we build a new dataset of real block towers in dynamic scenes, and evaluate the model's generalization power to real videos.

We further present ablation studies to understand how each source of information contributes to the final performance. We also conduct human behavioral experiments to contrast the performance of the model with humans. In a `which block is heavier' experiment, our model performs comparably to humans. 

Our contributions in this paper are three-fold. First, we propose the problem of physical primitive decomposition---learning a compact, disentangled object representation in terms of physical primitives. Second, we present a novel learning paradigm that learns to characterize shapes in physical primitives to explain both their geometry and physics. Third, we demonstrate that our system can achieve good performance on both synthetic and real data.
\section{Related Work}

\myparagraph{Primitive-Based 3D Representations.}
Early attempts on modeling 3D shapes with primitives include decomposing them into blocks~\cite{roberts1963machine}, generalized cylinders~\cite{binford1971visual}, and geons~\cite{Biederman1987Recognition}. This idea has been constantly revisited throughout the development of computer vision~\cite{Gupta2010Blocks,van2015part,Attene2006Hierarchical}. To name a few, Gupta~\etal~\cite{Gupta2010Blocks} modeled scenes as qualitative blocks, and van den Hengel~\etal~\cite{van2015part} as Lego blocks. More recently, Tulsaini~\etal~\cite{Tulsiani2017Learning} combined the new and the old---using deep convolutional network to generate primitives of a given 3D shape; later, Zou~\etal proposed 3D-PRNN~\cite{Zou20173D}, enhancing the flexibility of the system by leveraging modern advancement in recurrent generative models~\cite{van2016pixel}.

Primitive-based representations have profound impact that goes far beyond the field of computer vision. Scientists have employed this representation for user-interactive design~\cite{igarashi1999teddy} and for teaching robots to grasp objects~\cite{miller2003automatic}. In the field of computer graphics, the idea of modeling shapes as primitives or parts has also been extensively explored~\cite{zheng2014recurring,yumer2012co,li2011globfit,Kalogerakis2012Probabilistic,Kim2013Learning,Attene2006Hierarchical}. Researchers have used the part-based representation for single-image shape reconstruction~\cite{huang2015single}, shape completion~\cite{schnabel2009completion}, and probabilistic shape synthesis~\cite{Huang2015Analysis,li2017grass}. 

\myparagraph{Physical Shape and Scene Modeling}
Beyond object geometry, there have been growing interests in modeling physical object properties and scene dynamics. The computer vision community has put major efforts in building rich and sizable databases. ShapeNet-Sem~\cite{Savva2015Semantically} is a collection of object shapes with material and physics annotations within the web-scale shape repository ShapeNet~\cite{Chang2015Shapenet:}. Material in Context Database (MINC)~\cite{Bell2015Material} is a gigantic dataset of materials in the wild, associating patches in real-world images with 23 materials. 

Research on physical object modeling dates back to the study of ``functional parts''~\cite{Rivlin1995Recognition,Tenenbaum1994Functional,Gibson1977Affordances}. The field of learning object physics and scene dynamics has prospered in the past few years~\cite{Lerer2016Learning,Agrawal2016Learning,Jia20153D,Battaglia2013Simulation,Zhao2013Scene,Mottaghi2016What,pham2015towards,brubaker2010physics,soo2016force,kim2017data,li2017visual}. Among them, there are a few papers that explicitly build physical object representations~\cite{Mottaghi2016What,Wu2016Physics,Wu2015Galileo:,vda,uai}. Though they also focus on understanding object physics~\cite{Wu2016Physics,Wu2015Galileo:}, functionality~\cite{Zhu2015Understanding,yao2013discovering}, and affordances~\cite{koppula2014physically,grabner2011makes,zhu2014reasoning}, these approaches usually assume a homogeneous object with simple geometry. In our paper, we model an object using physical primitives for richer expressiveness and higher precision.
\section{Physical Primitive Decomposition}

\subsection{Problem Statement}

\begin{figure}[t]
\centering
\begin{subfigure}{0.45\linewidth}
    \centering
    \includegraphics[width=\linewidth]{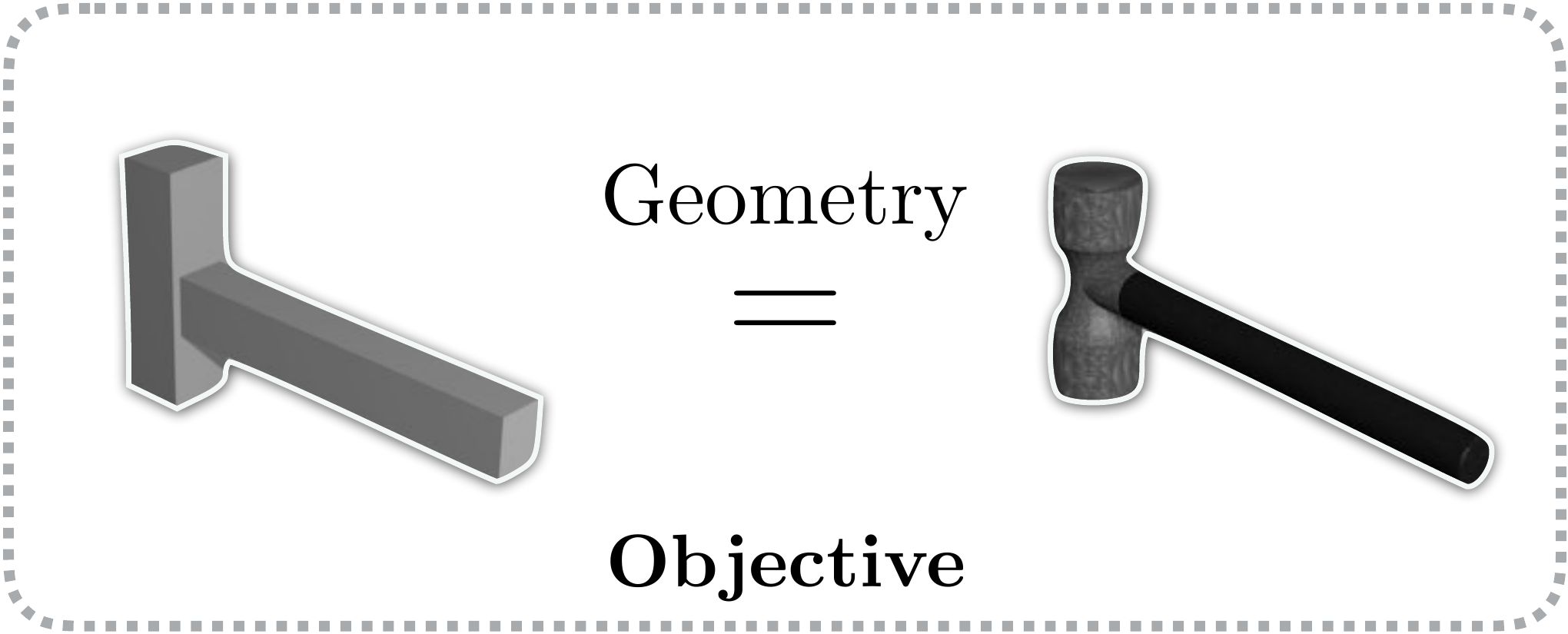}
    \caption{Primitive decomposition}
\end{subfigure}
\begin{subfigure}{0.45\linewidth}
    \centering
    \includegraphics[width=\linewidth]{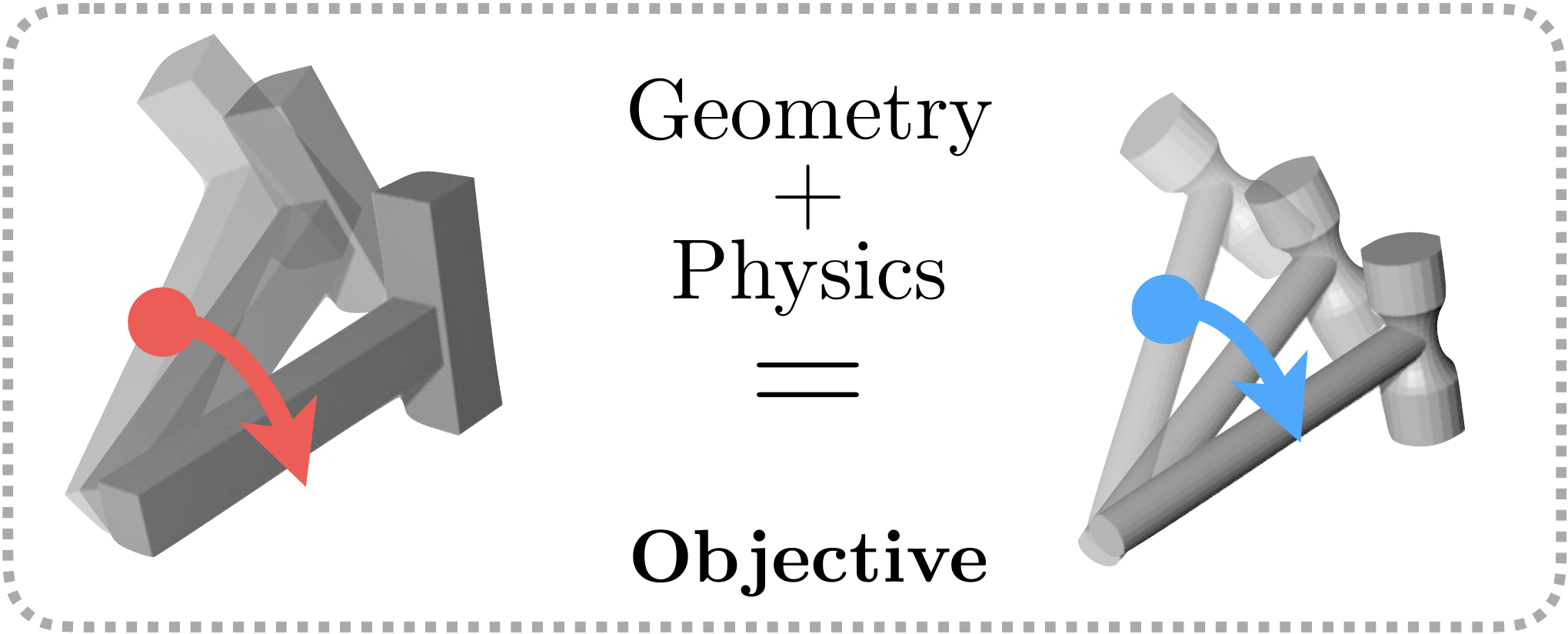}
    \caption{Physical primitive decomposition}
\end{subfigure}
\vspace{-8pt}
\caption{Primitive decomposition \textbf{(a)} and physical primitive decomposition \textbf{(b)}. Both tasks attempt to convert an object into a set of primitives yet with different purposes: the former problem targets at shape reconstruction, while the latter one aims to recover both geometric and physical properties.}
\label{fig:problem}
\vspace{-20pt}
\end{figure}

Both primitive decomposition and physical primitive decomposition attempt to approximate an object with primitives. We highlight their difference in \fig{fig:problem}.

\myparagraph{Primitive Decomposition}
As formulated in Tulsaini~\etal~\cite{Tulsiani2017Learning} and Zou~\etal~\cite{Zou20173D}, primitive decomposition aims to decompose an object $O$ into a set of simple transformed primitives $x = \{x_k\}$ so that these primitives can accurately approximate its geometry shape. This task can be seen as to minimize
\begin{equation}
    \mathcal{L}_\text{G}(x) = \mathcal{D}_\text{S} \big( \mathcal{S} \big( \underset{k}{\cup} x_k \big), \mathcal{S}(O) \big),
\end{equation}
where $\mathcal{S}(\cdot)$ denotes the geometry shape (\ie point cloud), and $\mathcal{D}_\text{S}(\cdot, \cdot)$ denotes the distance metric between shapes (\ie earth-mover's distance~\cite{Rubner2000earth}).

\myparagraph{Physical Primitive Decomposition}
In order to understand the functionality of object parts, we require the decomposed primitives $x = \{x_k\}$ to also approximate the physical behavior of object $O$. To this end, we extend the previous objective function with an additional physics term: 
\begin{equation}
    \mathcal{L}_\text{P}(x) = \sum_{p \in \mathcal{P}} \mathcal{D}_\text{T} \big( \mathcal{T}_p \big( \underset{k}{\cup} x_k \big), \mathcal{T}_p(O) \big),
\end{equation}
where $\mathcal{T}_p(\cdot)$ denotes the trajectory after physics interaction $p$, $\mathcal{D}_\text{T}(\cdot, \cdot)$ denotes the distance metric between trajectories (\ie mean squared error), and $\mathcal{P}$ denotes a predefined set of physics interactions. Therefore, the task of physical primitive decomposition is to minimize an overall objective function constraining both geometry and physics: $\mathcal{L}(x) = \mathcal{L}_\text{G}(x) + w \cdot \mathcal{L}_\text{P}(x)$, where $w$ is a weighting factor.

\begin{figure}[t]
\centering
\begin{subfigure}{0.49\linewidth}
    \centering
    \includegraphics[width=0.3\linewidth]{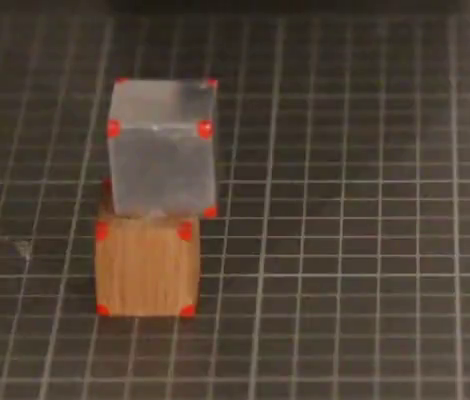}
    \includegraphics[width=0.3\linewidth]{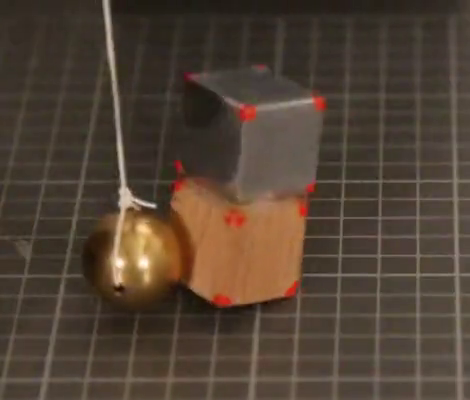}
    \includegraphics[width=0.3\linewidth]{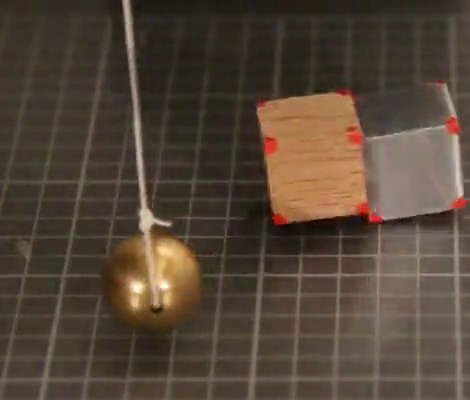}

    \includegraphics[width=0.3\linewidth]{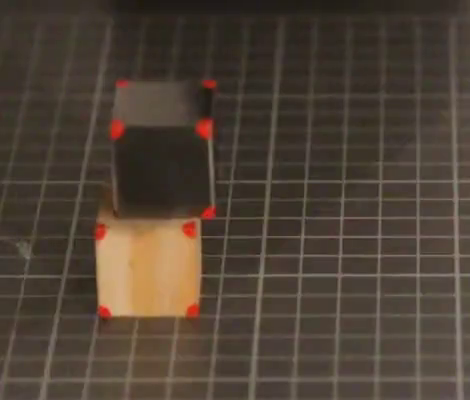}
    \includegraphics[width=0.3\linewidth]{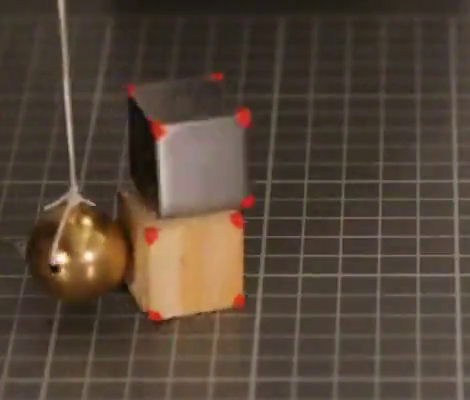}
    \includegraphics[width=0.3\linewidth]{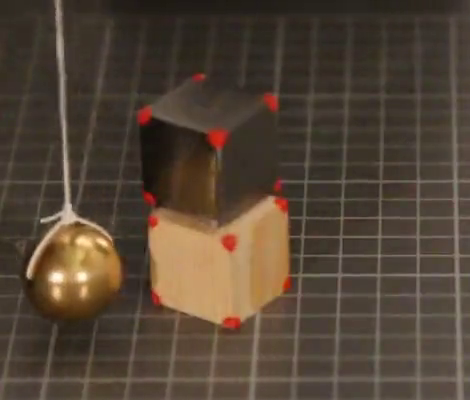}
    \caption{\centering \textbf{Above}: Aluminum and Wood; \newline \textbf{Below}: Iron and Wood.}
    \label{fig:challenges:a}
\end{subfigure}
\begin{subfigure}{0.49\linewidth}
    \centering
    \includegraphics[width=0.3\linewidth]{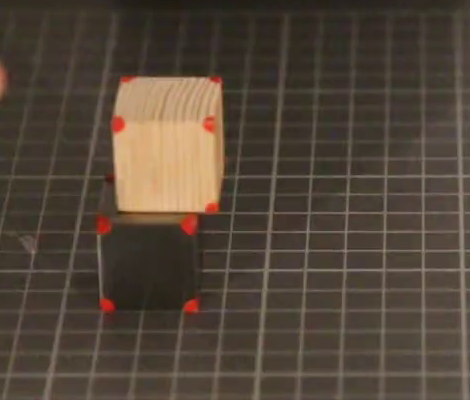}
    \includegraphics[width=0.3\linewidth]{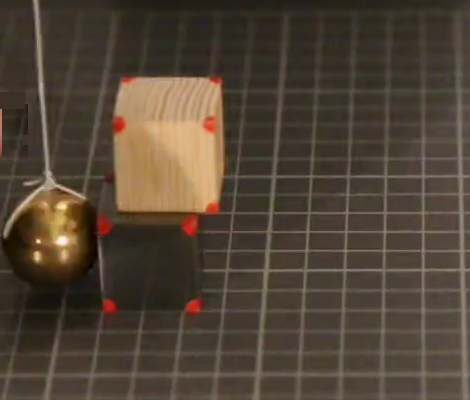}
    \includegraphics[width=0.3\linewidth]{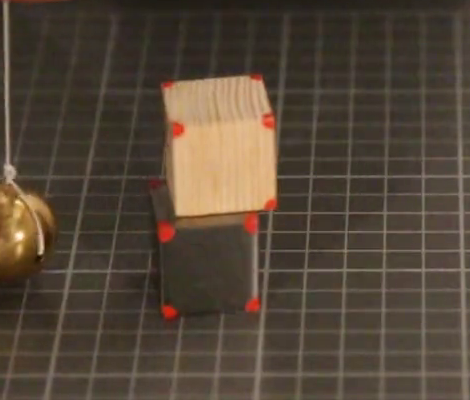}

    \includegraphics[width=0.3\linewidth]{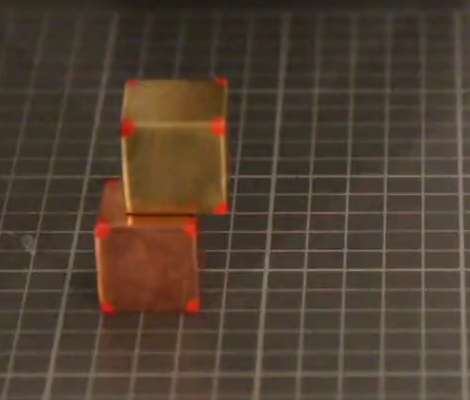}
    \includegraphics[width=0.3\linewidth]{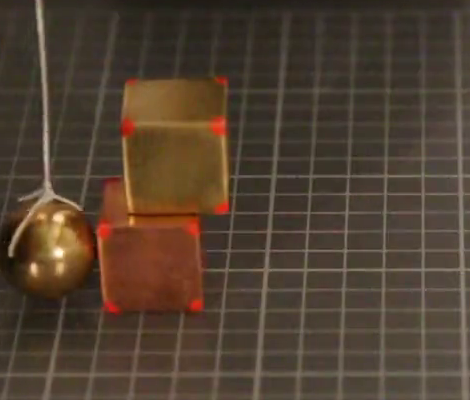}
    \includegraphics[width=0.3\linewidth]{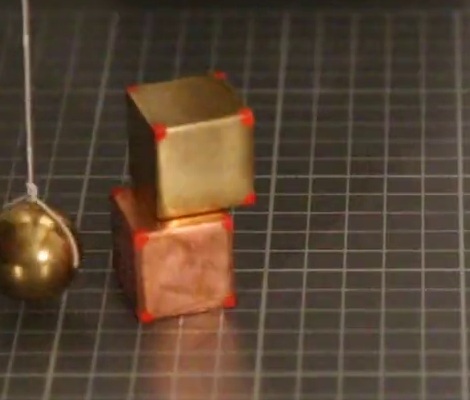}
    \caption{\centering \textbf{Above}: Wood and Iron; \newline \textbf{Below}: Two Coppers.}
    \label{fig:challenges:b}
\end{subfigure}
\vspace{-8pt}
\caption{Challenges of inferring physical parameters from visual and physical observations: objects with different physical parameters might have \textbf{(a)} similar visual appearance or \textbf{(b)} similar physics trajectory.}
\label{fig:challenges}
\vspace{-20pt}
\end{figure}

\subsection{Primitive-Based Representation}

We design a structured primitive-based object representation, which describes an object by listing all of its primitives with different attributes. For each primitive $x_k$, we record its size $x^\text{S}_k = (s_x, s_y, s_z)$, position in 3D space $x^\text{T}_k = (p_x, p_y, p_z)$, rotation in quaternion form $x^\text{R}_k = (q_w, q_x, q_y, q_z)$. Apart from these geometry information, we also track its physical properties: density $x^\text{D}_k$.

In our object representation, the shape parameters, $x^\text{S}_k$, $x^\text{T}_k$ and $x^\text{R}_k$, are vectors of continuous real values, whereas the density parameter $x^\text{D}_k$ is a discrete value. We discretize the density values into $N_\text{D} = 100$ slots, so that estimating density becomes a $N_\text{D}$-way classification. Discretization helps to deal with multi-modal density values. \fig{fig:challenges:a} shows that two parts with similar visual appearance may have very different physical parameters. In such cases, regression with an $\mathcal{L}_2$ loss will encourage the model to predict the \emph{mean} value of possible densities; in contrast, discretization allows it to give high probabilities to every possible density. We then figure out which candidate value is optimal from the trajectories. 
\section{Approach}

\begin{figure}[t]
\centering
\includegraphics[width=\linewidth]{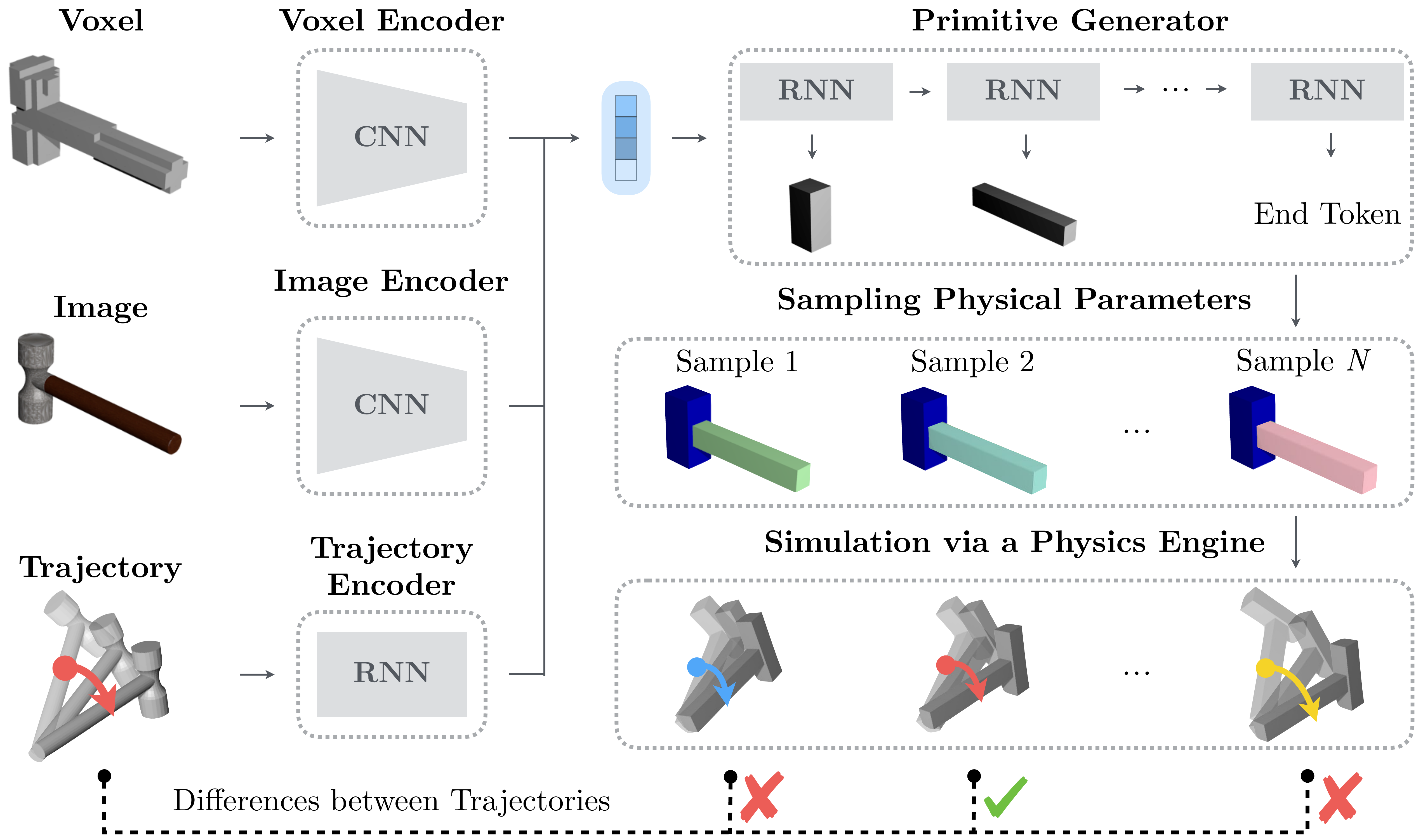}
\vspace{-15pt}
\caption{Overview of our \model model.}
\label{fig:model}
\vspace{-17pt}
\end{figure}

In this section, we discuss our approach to the problem of physical primitive decomposition (\model). We present an overview of our framework in \fig{fig:model}.

\subsection{Overview}

Inferring physical parameters from solely visual or physical observation is highly challenging. This is because two objects with different physical parameters might have similar visual appearance (\fig{fig:challenges:a}) or have similar physics trajectories (\fig{fig:challenges:b}). Therefore, our model takes both types of observations as input:
\begin{enumerate}
    \item \textbf{Visual Observation}. We take a voxelized shape and an image as our input because they can provide us with valuable visual information. Voxels help us recover object geometry, and images contain texture information of object materials. Note that, even with voxels as input, it is still highly nontrivial to infer geometric parameters: the model needs to learn to segment 3D parts within the object — an unsolved problem by itself~\cite{Tulsiani2017Learning}.
    \item \textbf{Physics Observation}. In order to explain the physical behavior of an object, we also need to observe its response after some physics interactions. In this work, we choose to use 3D object trajectories rather than RGB (or RGB-D) videos. Its abstractness enables the model to transfer better from synthetic to real data, because synthetic and real videos can be starkly different; in contrast, it’s easy to generate synthetic 3D trajectories that look realistic.
\end{enumerate}
\vspace{-5pt}
Specifically, our network takes a voxel $V$, an image $I$, and $N_\text{T}$ object trajectories $\bm{T} = \{T_k\}$ as input. $V$ is a 3D binary voxelized grid, $I$ is a single RGB image, and $\bm{T}$ consists of several object trajectories $T_k$, each of which records the response to one specific physics interaction. Trajectory $T_k$ is a sequence of 3D object pose $(p_x, p_y, p_z, q_w, q_x, q_y, q_z)$, where $(p_x, p_y, p_z)$ denotes the object's center position and quaternion $(q_w, q_x, q_y, q_z)$ denotes its rotation at each time step.

After receiving the inputs, our network encodes voxel, image and trajectory with separate encoders, and sequentially predicts primitives using a recurrent primitive generator. For each primitive, the network predicts its geometry shape (\ie scale, translation and rotation) and physical property (\ie density). More details of our model can be found in the supplementary material.

\myparagraph{Voxel Encoder}
For input voxel $V$, we employ a 3D volumetric convolutional network to encode the 3D shape information into a voxel feature $f_\text{V}$.

\myparagraph{Image Encoder}
For input image $I$, we pass it into the ResNet-18~\cite{He2015Deep} encoder to obtain an image feature $f_\text{I}$. We refer the readers to He~\etal~\cite{He2015Deep} for details.

\myparagraph{Trajectory Encoder}
For input trajectories $\bm{T}$, we encode each trajectory $T_k$ into a low-dimensional feature vector $h_k$ with a separate bi-directional recurrent neural network. Specifically, we feed the trajectory sequence, $T_k$, and also the same trajectory sequence in reverse order, $T_k^\text{reverse}$, into two encoding RNNs, to obtain two final hidden states: $h^\rightarrow_k = \text{encode}^\rightarrow_k(T_k)$ and $h^\leftarrow_k = \text{encode}^\leftarrow_k(T_k^\text{reverse})$. We take $[h_k^\rightarrow; h_k^\leftarrow]$ as the feature vector $h_k$. Finally, we concatenate the features of each trajectory, $\{h_k \mid k = 1, 2, \ldots, N_\text{T}\}$, and project it into a low-dimensional trajectory feature $f_\text{T}$ with a fully-connected layer.

\myparagraph{Primitive Generator}
We concatenate the voxel feature $f_\text{V}$, image feature $f_\text{I}$ and trajectory feature $f_\text{T}$ together as $\hat{f} = [f_\text{V}; f_\text{I}; f_\text{T}]$, and map it to a low-dimensional feature $f$ using a fully-connected layer. We predict the set of physical primitives $\{x_k\}$ sequentially by a recurrent generator.

At each time step $k$, we feed the previous generated primitive $x_{k-1}$ and the feature vector $f$ in as input, and we receive one hidden vector $h_k$ as output. Then, we compute the new primitive $x_k = (x^\text{D}_k, x^\text{S}_k, x^\text{T}_k, x^\text{R}_k)$ as 
\abovedisplayskip=2pt
\belowdisplayskip=2pt
\begin{equation}
    \begin{aligned}
        x^\text{D}_k &= \text{softmax}(W_\text{D} \times h_k + b_\text{D}), \quad x^\text{S}_k = \text{sigmoid}(W_\text{S} \times h_k + b_\text{S}) \times C_\text{S}, \\
        x^\text{T}_k &= \tanh(W_\text{T} \times h_k + b_\text{T}) \times C_\text{T}, \quad x^\text{R}_k = \frac{W_\text{R} \times h_k + b_\text{R}}{\max(\norm{W_\text{R} \times h_k + b_\text{R}}_2, \epsilon)},
    \end{aligned}
    \label{eqn:primitive}
\end{equation}
where $C_\text{S}$ and $C_\text{T}$ are scaling factors, and $\epsilon = 10^{-12}$ is a small constant for numerical stability. \eqn{eqn:primitive} guarantees that $x^\text{S}_k$ is in the range of $[0, C_\text{S}]$, $x^\text{T}_k$ is in the range of $[-C_\text{T}, C_\text{T}]$, and $\norm{x^\text{R}_k}_2$ is $1$ (if ignoring $\epsilon$), which ensures that $x_k$ will always be a valid primitive. In our experiments, we set $C_\text{S} = C_\text{T} = 0.5$, since we normalize all objects so that they can fit in unit cubes. Also note that, $x^\text{D}_k$ is an $(N_\text{D} + 2)$-dimensional vector, where the first $N_\text{D}$ dimensions indicate different density values and the last two indicate the ``start token" and ``end token".

\myparagraph{Sampling and Simulating with the Physics Engine}
During testing time, we treat the predicted $x^\text{D}_k$ as a multinomial distribution, and we sample multiple possible predictions from it. For each sample, we use its physical parameters to simulate the trajectory with a physics engine. Finally, we select the one whose simulated trajectory is closest to the observed trajectory.

An alternative way to incorporate physics engine is to directly optimize our model over it. As most physics engines are not differentiable, we employ REINFORCE~\cite{Williams1992Simple} for optimization. Empirically, we observe that this reinforcement learning based method performs worse than sampling-based methods, possible due to the large variance of the approximate gradient signals. 

Simulating with a physics engine requires we know the force during testing. Such an assumption is essential to ensure the problem is well-posed: without knowing the force, we can only infer the relative part density, but not the actual values. Note that in many real-world applications such as robot manipulation, the external force is indeed available. 

\subsection{Loss Functions}

Let $x = (x_1, x_2, \ldots, x_n)$ and $\hat{x} = (\hat{x}_1, \hat{x}_2, \ldots, \hat{x}_m)$ be the predicted and ground-truth physical primitives, respectively. Our loss function consists of two terms, geometry loss $\mathcal{L}_\text{G}$ and physics loss $\mathcal{L}_\text{D}$:
\abovedisplayshortskip=0pt
\belowdisplayshortskip=0pt
\abovedisplayskip=0pt
\belowdisplayskip=0pt
\begin{align}
    \mathcal{L}_\text{G}(x, \hat{x}) &= \sum_k \left( \omega_\text{S} \cdot \norm{x^\text{S}_k - \hat{x}^\text{S}_k}_1 + \omega_\text{T} \cdot \norm{x^\text{T}_k - \hat{x}^\text{T}_k}_1 + \omega_\text{R} \cdot \norm{x^\text{R}_k - \hat{x}^\text{R}_k}_1 \right),
    \label{eqn:loss:g} \\
    \mathcal{L}_\text{P}(x, \hat{x}) &= - \sum_k \sum_i \hat{x}^\text{D}_k (i) \cdot \log x^\text{D}_k(i),
\label{eqn:loss:p}
\end{align}
where $\omega_\text{S}$, $\omega_\text{T}$ and $\omega_\text{R}$ are weighting factors, which are set to 1's because $x^\text{S}$, $x^\text{T}$ and $x^\text{R}$ are of the same magnitude ($10^{-1}$) in our datasets. Integrating \eqn{eqn:loss:g} and \eqn{eqn:loss:p}, we define the overall loss function as $\mathcal{L}(x, \hat{x}) = \mathcal{L}_\text{G}(x, \hat{x}) + w \cdot \mathcal{L}_\text{P}(x, \hat{x})$, where $w$ is set to ensure that $\mathcal{L}_G$ and $\mathcal{L}_P$ are of the same magnitude.

\myparagraph{Part Associations}
In our formulation, object parts (physical primitives) follow a pre-defined order (\eg, from bottom to top), and our model is encouraged to learn to predict the primitives in the same order.
\section{Experiments}
\label{sec:exp}

We evaluate our \model model on three diverse settings: synthetic block towers where blocks are of various materials and shapes; synthetic tools with more complex geometry shapes; and real videos of block towers to demonstrate our transferability to real-world scenario.

\subsection{Decomposing Block Towers}
\label{sec:exp:blocks}

We start with decomposing block towers (stacks of blocks).

\begin{figure}[!t]
\centering
\includegraphics[width=0.95\linewidth]{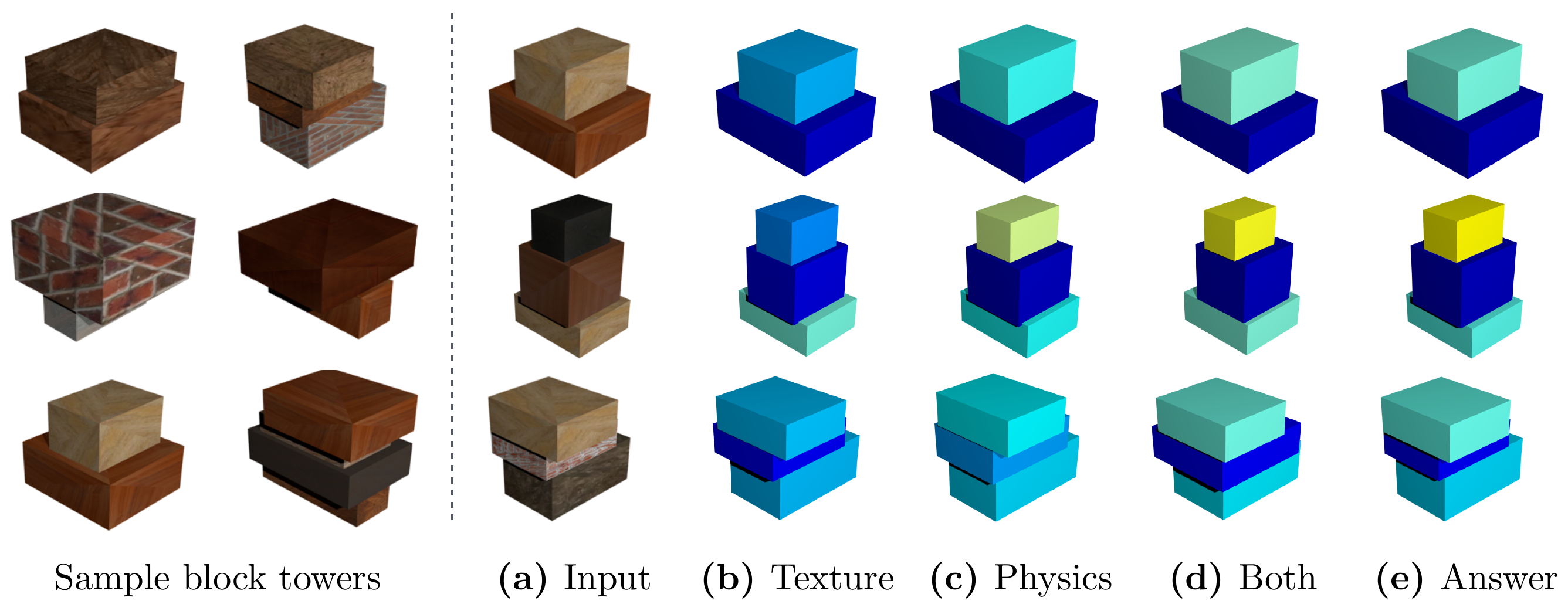}
\includegraphics[width=0.95\linewidth]{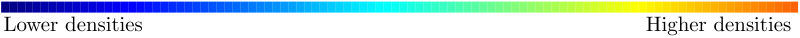}
\vspace{-8pt}
\caption{Sample objects in our block towers dataset (left) and qualitative results of our model with different combinations of observations as input (right).}
\label{fig:blocks_data}
\vspace{-8pt}
\end{figure}

\myparagraph{Block Towers} 
We build the block towers by stacking variable number of blocks (2-5 in our experiments) together. We first sample the size of each block and then compute the center position of blocks from bottom to top. For the $k$\textsuperscript{th} block, we denote the size as $(w_k, h_k, d_k)$, and its center $(x_k, y_k, z_k)$ is sampled and computed by $x_k \sim \mathcal{N}(x_{k-1}, w_{k-1}/4)$, $y_k \sim \mathcal{N}(y_{k-1}, h_{k-1}/4)$, and $z_k = z_{k-1} + (d_{k-1} + d_k) / 2$, where $\mathcal{N}(\mu, \sigma)$ is a normal distribution with mean $\mu$ and standard deviation $\sigma$. We illustrate some constructed block towers in \fig{fig:blocks_data}. We perform the exact voxelization with grid size of 32$\times$32$\times$32 by binvox, a 3D mesh voxelizer~\cite{nooruddin03binvox}.

\begin{table}[!t]
\setlength{\tabcolsep}{8pt}
\centering\small
\begin{tabular}{cccccc}
    \toprule
    Material & Wood & Brick & Stone & Ceramic & Metal\\
    \midrule
    Density & $[1, 10]$ & $[11, 20]$ & $[21, 30]$
    & $[31, 60]$ & $[21, 35] \cup [71, 100]$\\
    \bottomrule
\end{tabular}
\vspace{3pt}
\caption{Materials and their real-world density values (unit: $\times 10^2 \cdot \text{kg}/\text{m}^3$). Objects made of similar materials (different types of metals) may have different physical properties, while different materials (\ie, stone and metal) may have same physical properties.}
\label{tbl:material}
\vspace{-28pt}
\end{table}

\myparagraph{Materials}
In our experiments, we use five different materials, and follow their real-world densities with minor modifications. The materials and the ranges of their densities are listed in \tbl{tbl:material}. For each block in the block towers, we first assign it to one of the five materials, and then uniformly sample its density from possible values of its material. We generate 8 configurations for each block tower.

\myparagraph{Textures}
We obtain the textures for materials by cropping the center portion of images from the MINC dataset~\cite{Bell2015Material}. We show sample images rendered with material textures in \fig{fig:blocks_data}. Since we render the textures only with respect to the material, the images rendered do not provide any information about density.

\myparagraph{Physics Interactions}
We place the block towers at the origin and perform four physics interactions to obtain the object trajectories ($N_\text{T} = 4$). In detail, we exert a force with the magnitude of $10^5$ on the block tower from four pre-defined positions $\{(\pm 1, -1, \pm 1)\}$. We simulate each physics interaction for 256 time steps using the Bullet Physics Engine~\cite{Coumans2010Bullet}. To ensure simulation accuracy, we set the time step for simulation to $1/300$s.

\begin{table}[t]
\setlength{\tabcolsep}{2.9pt}
\centering\small
\begin{tabular}{lccC{1cm}C{1cm}C{1cm}C{1cm}c}
    \toprule
    \multirow{3}{*}{Methods} & \multicolumn{2}{c}{Observations} & \multicolumn{4}{c}{Density} & Trajectory \\
    \cmidrule(lr){2-3}\cmidrule(lr){4-7}\cmidrule{8-8}
    & \multirow{2}{*}{Texture} & \multirow{2}{*}{Physics} & \multicolumn{3}{c}{Accuracy} & \multirow{2}{*}{RMSE} & \multirow{2}{*}{MAE} \\
    \cmidrule(lr){4-6}
    & & & Top 1 & Top 5 & Top 10 \\
    \midrule
    Frequent & \textbf{--} & \textbf{--} & 2.0 & 9.7 & 13.4 & 25.4 & 74.4 \\
    Nearest & \textbf{--} & \textbf{+} & 1.9 & 7.9 & 12.4 & 41.1 & 91.0 \\
    Oracle & \textbf{+} & \textbf{--} & 6.9 & 35.7 & 72.0 & 18.5 & 51.3 \\
    \midrule
    \model (no trajectory) & \textbf{+} & \textbf{--} & 7.2 & 35.2 & 69.5 & 19.0 & 51.7 \\
    \model (no image)  & \textbf{--} & \textbf{+} & 7.1 & 31.0 & 50.8 & 16.7 & 36.4 \\
    \model (no voxels)  & \textbf{+} & \textbf{+} & 15.9 & 56.3 & 82.4 & 10.3 & 29.9 \\
    \midrule
    \model (RGB-D) & \textbf{+} & \textbf{+} & 11.6 & 50.5 & 79.5 & 12.8 & 30.2 \\
    \model (full) & \textbf{+} & \textbf{+} & 16.1 & 56.4 & 82.5 & 9.9 & 21.0 \\
    \model (full)+Sample & \textbf{+} & \textbf{+}&  \textbf{18.2} & \textbf{59.7} & \textbf{84.0} & \textbf{8.8} & \textbf{13.9}\\
    \bottomrule
\end{tabular}
\vspace{3pt}
\caption{Quantitative results of physical parameter estimation on block towers. Combining appearance with physics does help our model to achieve better estimation on physical parameters, and our model performs significantly better than all other baselines.}
\label{tbl:blocks_result}
\vspace{-32pt}
\end{table}

\myparagraph{Metrics}
We evaluate the performance of shape reconstruction by the $\text{F}_1$ score between the prediction and ground truth: each primitive in prediction is labeled as a true positive if its intersection over union (IoU) with a ground-truth primitive is greater than 0.5. For physics estimation, we employ two types of metrics, i) density measures: top-$k$ accuracy ($k \in \{1, 5, 10\}$) and root-mean-square error (RMSE) and ii) trajectory measure: mean-absolute error (MAE) between simulated trajectory (using predicted the physical parameters) and ground-truth trajectory.

\myparagraph{Methods}
We evaluate our model with different combinations of observations as input: i) texture only (\ie, no trajectory, by setting $f_\text{T} = 0$), ii) physics only (\ie, no image, by setting $f_\text{I} = 0$), iii) both texture and physics but without the voxelized shape, iv) both texture and physics but with replacing the 3D trajectory with a raw depth video, v) full data in our original setup (image, voxels, and trajectory). We also compare our model with several baselines: i) predicting the most frequent density in the training set (\emph{Frequent}), ii) nearest neighbor retrieval from the training set (\emph{Nearest}), and iii) knowing the ground-truth material and guessing within its density value range (\emph{Oracle}). While all these baselines assume perfect shape reconstruction, our model learns to decompose the shape.

\myparagraph{Results}
For the shape reconstruction, our model achieves 97.5 in terms of F1 score. For the physics estimation, we present quantitative results of our model with different observations as input in \tbl{tbl:blocks_result}. We compare our model with an oracle that infers material properties from appearance while assuming ground-truth reconstruction. It gives \emph{upper-bound} performance of methods that rely on only appearance cues. Experiments suggest that appearance alone is not sufficient for density estimation. From \tbl{tbl:blocks_result}, we observe that combining appearance with physics performs well on physical parameter estimation, which is because the object trajectories can provide crucial additional information about the density distribution (\ie moment of inertia). Also, all input modalities and sampling contribute to the model's final performance.

\begin{wraptable}{r}{.55\linewidth}
\vspace{-22pt}
\setlength{\tabcolsep}{2.5pt}
\centering\small
\begin{tabular}{lcccc}
    \toprule
     & 1$\times$ & 8$\times$ & 64$\times$ & 512$\times$\\
    \midrule
    Sample Phys.+Shape & 142.2 & 87.1 & 70.8 & 58.7 \\
    Sample Phys. & 89.7 & 60.1 & 38.7 & 22.7 \\
    \model (ours) & \textbf{21.0} & \textbf{15.1} & \textbf{13.9} & \textbf{13.2} \\
    \bottomrule
\end{tabular}
\vspace{-10pt}
\caption{Comparison between our model and a physics engine based sampling baseline}
\label{tbl:sampling_comparison}
\vspace{-27pt}
\end{wraptable}

We have also implemented a physics engine--based sampling baseline: sampling the shape and physical parameters for each primitive, using a physics engine for simulation, and selecting the one whose trajectory is closest to the observation. We also compare with a stronger baseline where we only sample physics, assuming ground-truth shape is known. \tbl{tbl:sampling_comparison} shows our model works better and is more efficient: the neural nets have learned an informative prior that greatly reduces the need of sampling at test time.

\subsection{Decomposing Tools}
\label{sec:exp:tools}

We then demonstrate the practical applicability of our model by decomposing synthetic real-world tools.

\begin{figure}[!t]
\centering
\includegraphics[width=0.95\linewidth]{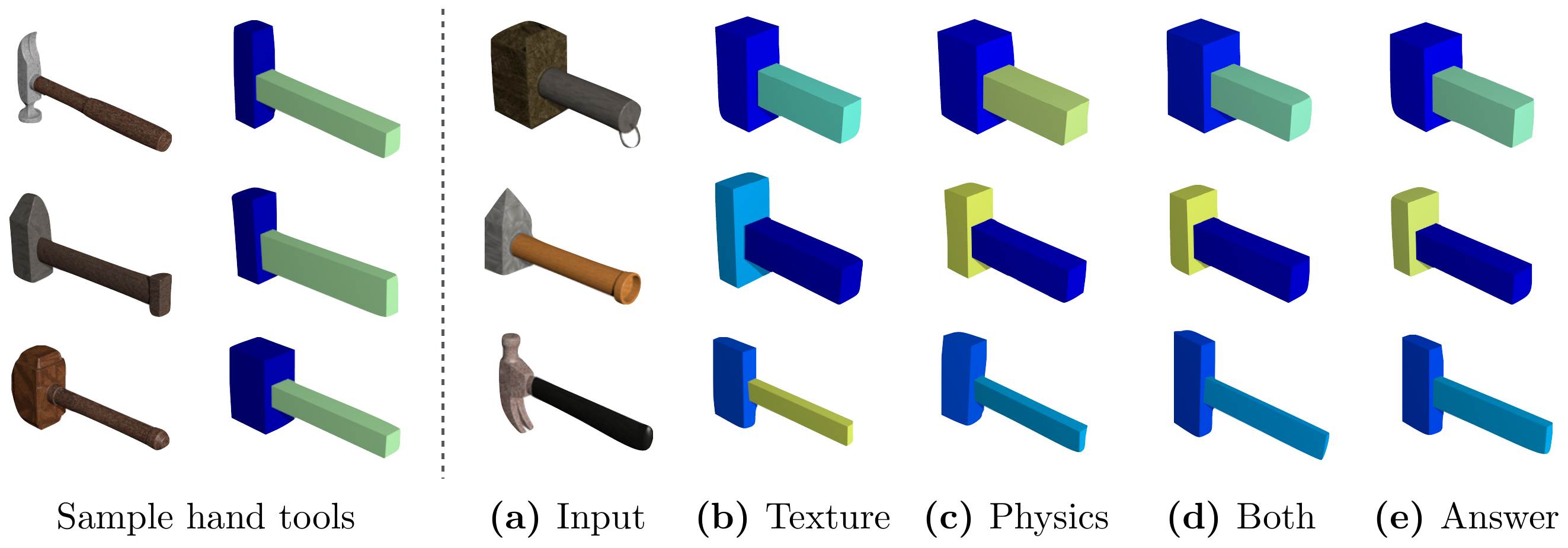}
\includegraphics[width=0.95\linewidth]{fig/colorbar.pdf}
\vspace{-8pt}
\caption{Sample objects in synthetic tools dataset (left) and qualitative results of our model with different combinations of observations as input (right).}
\label{fig:tools_data}
\vspace{-20pt}
\end{figure}

\myparagraph{Tools}
Because of the absence of tool data in the ShapeNet Core~\cite{Chang2015Shapenet:} dataset, we download the tools from 3D Warehouse\footnote{\url{https://3dwarehouse.sketchup.com}} and manually remove all unrelated models. In total, there are 204 valid tools, and we use Blender to remesh and clean up these tools to fix the issues with missing faces and normals. Following Chang~\etal~\cite{Chang2015Shapenet:}, we perform PCA on the point clouds and align models by their PCA axes. Sample tools in our dataset are shown in \fig{fig:tools_data}.

\begin{table}[t]
\setlength{\tabcolsep}{2.9pt}
\centering\small
\begin{tabular}{lccC{1cm}C{1cm}C{1cm}C{1cm}c}
    \toprule
    \multirow{3}{*}{Methods} & \multicolumn{2}{c}{Observations} & \multicolumn{4}{c}{Density} & Trajectory \\
    \cmidrule(lr){2-3}\cmidrule(lr){4-7}\cmidrule{8-8}
    & \multirow{2}{*}{Texture} & \multirow{2}{*}{Physics} & \multicolumn{3}{c}{Accuracy} & \multirow{2}{*}{RMSE} & \multirow{2}{*}{MAE} \\
    \cmidrule(lr){4-6}
    & & & Top 1 & Top 5 & Top 10 \\
    \midrule
    Frequent & \textbf{--} & \textbf{--} & 2.5 & 10.2 & 13.6 & 25.9 & 348.2 \\
    Nearest & \textbf{--} & \textbf{+} & 2.9 & 8.3 & 12.4 & 25.8 & 329.7 \\
    Oracle & \textbf{+} & \textbf{--} & 7.4 & 35.2 & 72.0 & 19.1 & 185.8 \\
    \midrule
    \model (no trajectory) & \textbf{+} & \textbf{--} & 7.7 & 36.4 & 71.1 & 16.8 & 206.8 \\
    \model (no image) & \textbf{--} & \textbf{+} & 15.0 & 56.3 & 80.2 & 5.9 & 143.6 \\
    \cmidrule{1-8}
    \model (full) & \textbf{+} & \textbf{+} & 35.7 & \textbf{85.2} & 95.8 & 2.6 & 103.6 \\
    \model (full)+Sample & \textbf{+} & \textbf{+} & \textbf{38.3} & 85.0 & \textbf{96.1} & \textbf{2.5} & \textbf{74.4}\\
    \bottomrule
\end{tabular}
\vspace{3pt}
\caption{Quantitative results of physical parameter estimation on tools. Combining visual appearance with physics observations helps our model to perform much better on physical parameter estimation, and compared to all other baselines, our model performs significantly better on this dataset.}
\label{tbl:tools_result}
\vspace{-28pt}
\end{table}

\myparagraph{Primitives}
Similar to Zou~\etal~\cite{Zou20173D}, we first use the energy-based optimization to fit the primitives from the point clouds, and then, we assign each vertex to its nearest primitive and refine each primitive with the minimum oriented bounding box of vertices assigned to it.

\myparagraph{Other Setups}
We make use of the same set of materials and densities as in \tbl{tbl:material} and the same textures for materials as described in \sect{sec:exp:blocks}. Sample images rendered with textures are shown in \fig{fig:tools_data}. As for physics interactions, we follow the same scenario configurations as in \sect{sec:exp:blocks}.

\myparagraph{Training Details}
Because the size of synthetic tools dataset is rather limited, we first pre-train our \model model on the block towers and then finetune it on the synthetic tools. For the block towers used for pre-training, we fix the number of blocks to 2 and introduce small random noises and rotations to each block to fill the gap between block towers and synthetic tools.

\myparagraph{Results}
For the shape reconstruction, our model achieves 85.9 in terms of F1 score. For the physics estimation, we present quantitative results in \tbl{tbl:tools_result}. The shape reconstruction is not as good as that of the block towers dataset because the synthetic tools are more complicated, and the orientations might introduce some ambiguity (there might exist multiple bounding boxes with different rotations for the same part of object). The physics estimation performance is better since the number of primitives in our synthetic tools dataset is very small ($\leq$2 in general). We also show some qualitative results in \fig{fig:tools_data}.

\subsection{Decomposing Real Objects}

We look into real objects to evaluate the generalization ability of our model.

\myparagraph{Real-World Block Towers}
We purchase totally ten sets of blocks with different materials (\ie pine, steel, aluminum and copper) from Amazon, and construct a dataset of real-world block towers. Our dataset contains 16 block towers with different configurations: 8 with two blocks, 4 with three blocks, and another 4 with four blocks.

\begin{figure}[t]
\centering
\begin{subfigure}{0.15\linewidth}
    \centering
    \includegraphics[width=\linewidth]{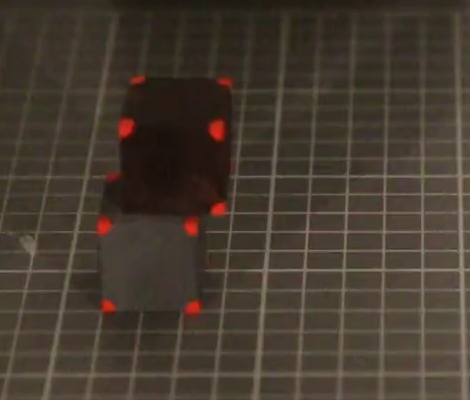}
    \includegraphics[width=\linewidth]{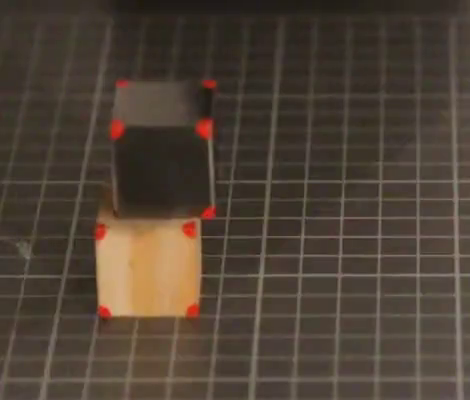}
    \includegraphics[width=\linewidth]{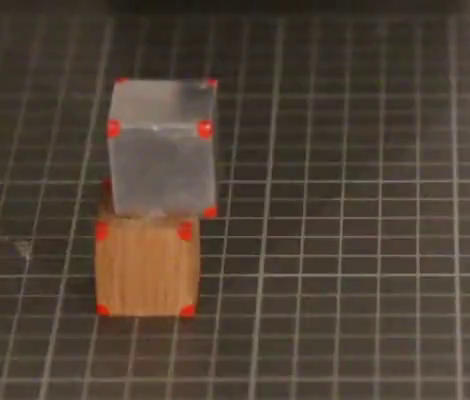}
    \caption{Frame $i_1$}
\end{subfigure}
\begin{subfigure}{0.15\linewidth}
    \centering
    \includegraphics[width=\linewidth]{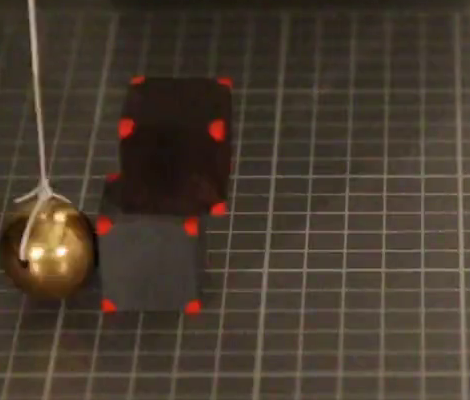}
    \includegraphics[width=\linewidth]{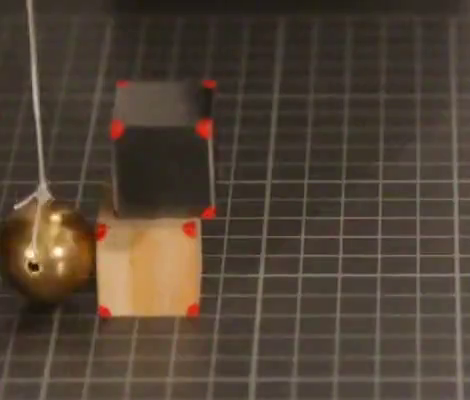}
    \includegraphics[width=\linewidth]{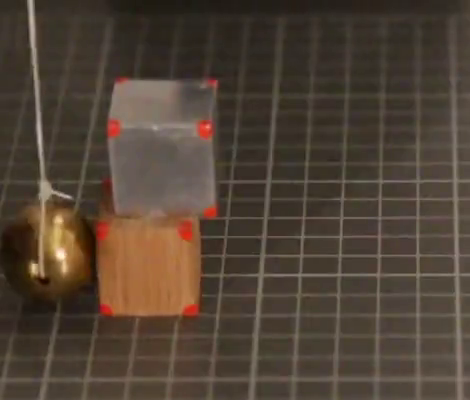}
    \caption{Frame $i_2$}
\end{subfigure}
\begin{subfigure}{0.15\linewidth}
    \centering
    \includegraphics[width=\linewidth]{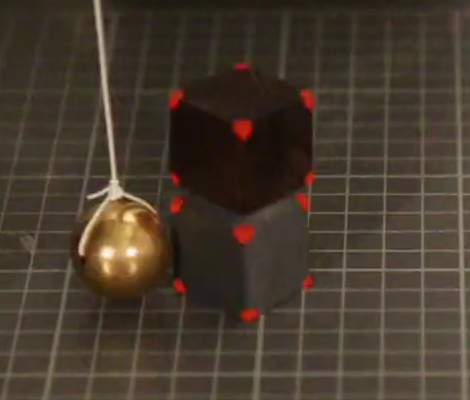}
    \includegraphics[width=\linewidth]{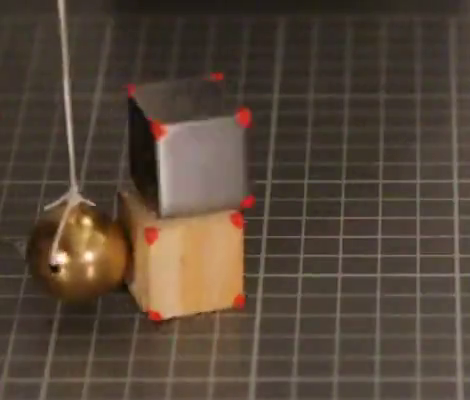}
    \includegraphics[width=\linewidth]{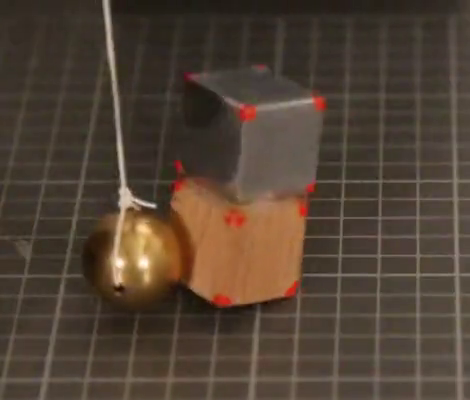}
    \caption{Frame $i_3$}
\end{subfigure}
\begin{subfigure}{0.15\linewidth}
    \centering
    \includegraphics[width=\linewidth]{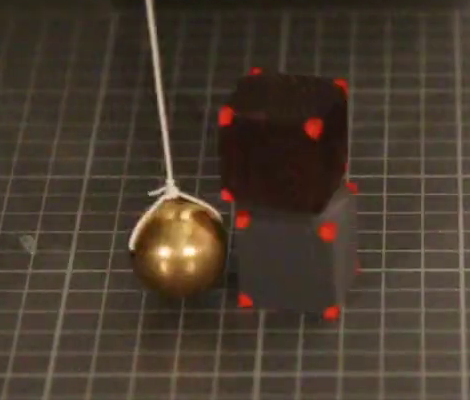}
    \includegraphics[width=\linewidth]{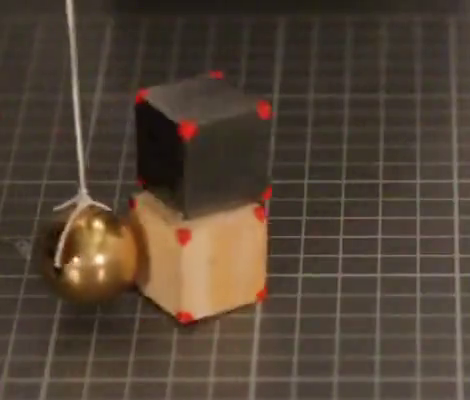}
    \includegraphics[width=\linewidth]{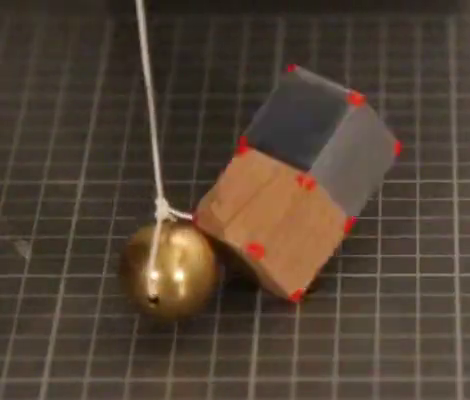}
    \caption{Frame $i_4$}
\end{subfigure}
\begin{subfigure}{0.15\linewidth}
    \centering
    \includegraphics[width=\linewidth]{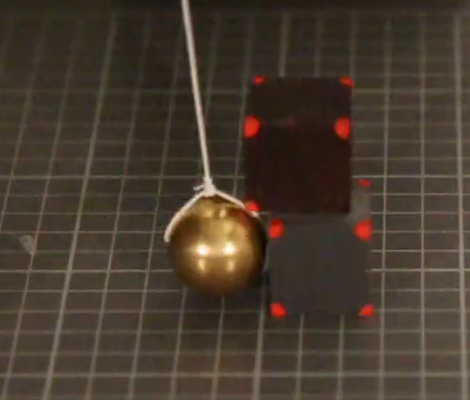}
    \includegraphics[width=\linewidth]{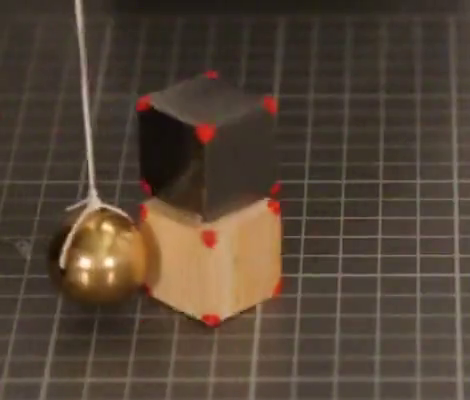}
    \includegraphics[width=\linewidth]{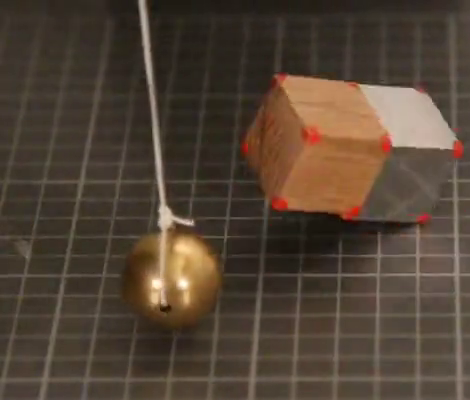}
    \caption{Frame $i_5$}
\end{subfigure}
\begin{subfigure}{0.15\linewidth}
    \centering
    \includegraphics[width=\linewidth]{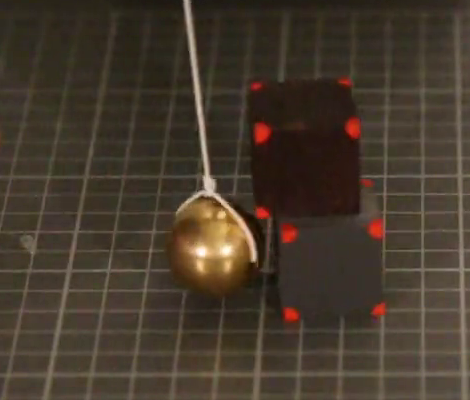}
    \includegraphics[width=\linewidth]{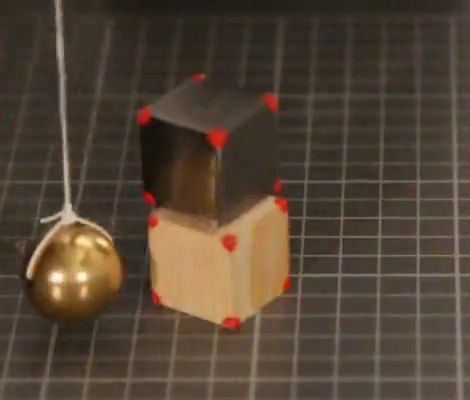}
    \includegraphics[width=\linewidth]{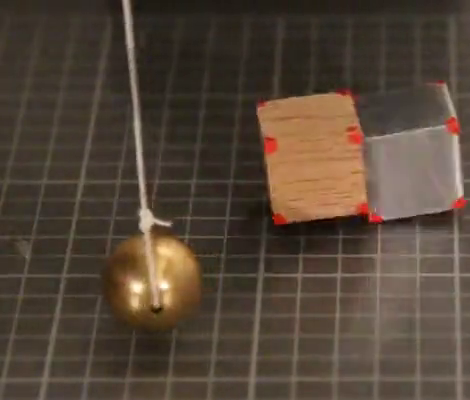}
    \caption{Frame $i_6$}
\end{subfigure}
\vspace{-8pt}
\caption{Objects and their physics trajectories in six sampled frames from our real-world block towers dataset. As in the last two rows, objects with similar visual appearances may have distinct physical properties that we can only distinguish from their behaviors in physical events.}
\label{fig:real_data}
\vspace{-8pt}
\end{figure}

\myparagraph{Physics Interaction}
The scenario is set up as follows: the block tower is placed at a specific position on the desk, and we use a copper ball (hang by a pendulum) to hit it. In \fig{fig:real_data}, we show some objects and their trajectories in our dataset.

\myparagraph{Video to 3D Trajectory}
On real-world data, the appearance of every frame in RGB video is used to extract a 3D trajectory. A major challenge is how to convert RGB videos into 3D trajectories. We employ the following approach:
\begin{enumerate}
    \item \textbf{Tracking 2D Keypoints}. For each frame, we first detect the 2D positions of object corners. For simplicity, we mark the object corners using red stickers and use a simple color filter to determine the corner positions. Then, we find the correspondence between the corner points from consecutive frames by solving the minimum-distance matching between two sets of points. After aligning the corner points in different frames, we obtain the 2D trajectories of these keypoints.
    \item \textbf{Reconstructing 3D Poses}. We annotate the 3D position for each corner point. Then, for each frame, we have 2D locations of keypoints and their corresponding 3D locations. Finally, we reconstruct the 3D object pose in each frame by solving the Perspective-n-Point between 2D and 3D locations using Levenberg-Marquardt algorithm~\cite{levenberg1944method,marquardt1963algorithm}.
\end{enumerate}

\begin{figure}[t]
\centering
\includegraphics[width=0.95\linewidth]{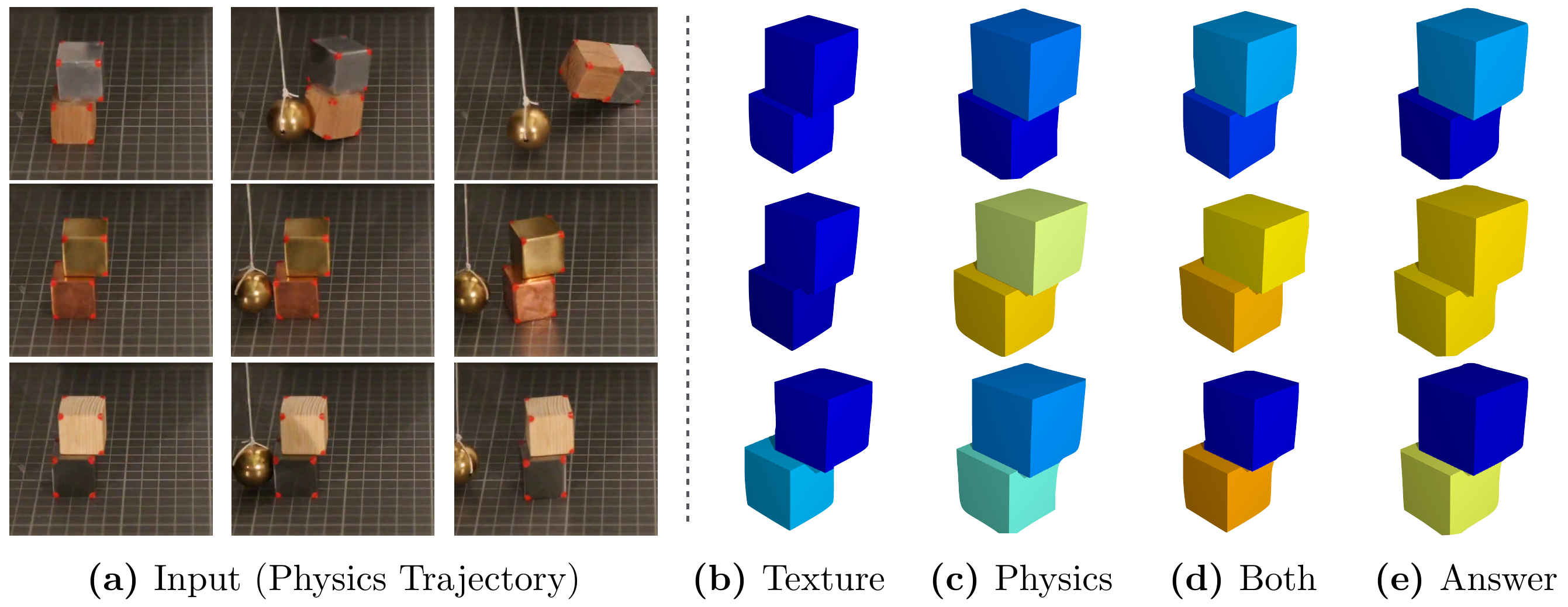}
\includegraphics[width=0.95\linewidth]{fig/colorbar.pdf}
\vspace{-8pt}
\caption{Qualitative results (on real-world block towers) of our model with different combinations of observations as input.}
\label{fig:real_result}
\vspace{-20pt}
\end{figure}

\myparagraph{Training Details}
We build a virtual physics environment, similar to our real-world setup, in the Bullet Physics Engine~\cite{Coumans2010Bullet}. We employ it to simulate physics interactions and generate a dataset of synthetic block towers to train our model.

\myparagraph{Results}
We show some qualitative results of our model with different observations as input in \fig{fig:real_result}. In the real-world setup, with only texture or physics information, our model cannot effectively predict the physical parameters because images and object trajectories are much noisier than those in synthetic dataset, while combining them together indeed helps it to predict much more accurate results. In terms of quantitative evaluation, our model (with both observations as input) achieves an RMSE value of 18.7 over the whole dataset and 10.1 over the block towers with two blocks (the RMSE value of random guessing is 40.8).
\section{Analysis}

To better understand our model, we present several analysis. The first three are conducted on synthetic block towers and the last one is on our real dataset.

\begin{figure}[t]
\centering
\includegraphics[width=\linewidth]{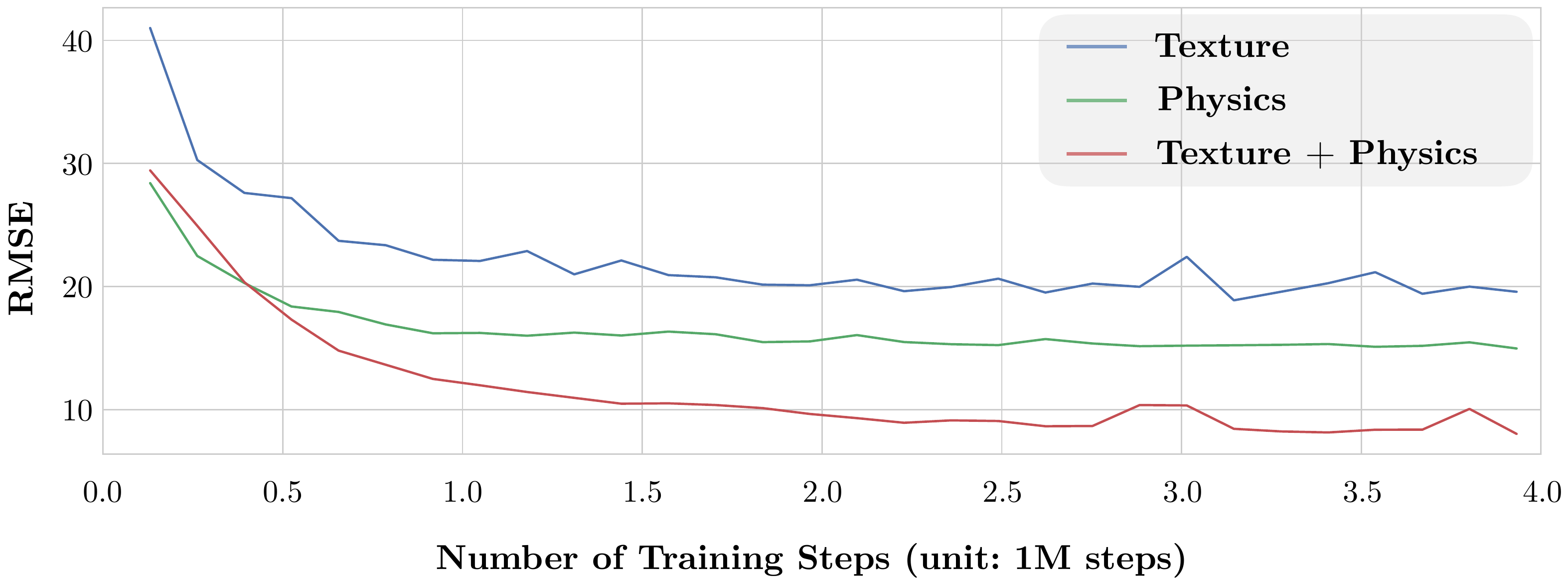}
\vspace{-18pt}
\caption{Learning curves with different observations as input. Our model learns much better and faster when both texture and physics supervisions are available.}
\label{fig:blocks_curve}
\vspace{-8pt}
\end{figure}

\myparagraph{Learning Speed with Different Supervisions}
We show the learning curves of our \model model with different supervision in \fig{fig:blocks_curve}. Model supervised by physics observation reaches the same level of performance of model with texture supervision using much fewer training steps (500K \vs 2M). Supervised by both observations, our \model model preserves the learning speed of the model with only physics supervision, and further improves its performance.

\begin{figure}[t]
\centering

\begin{subfigure}{0.32\linewidth}
    \centering
    \includegraphics[width=\linewidth]{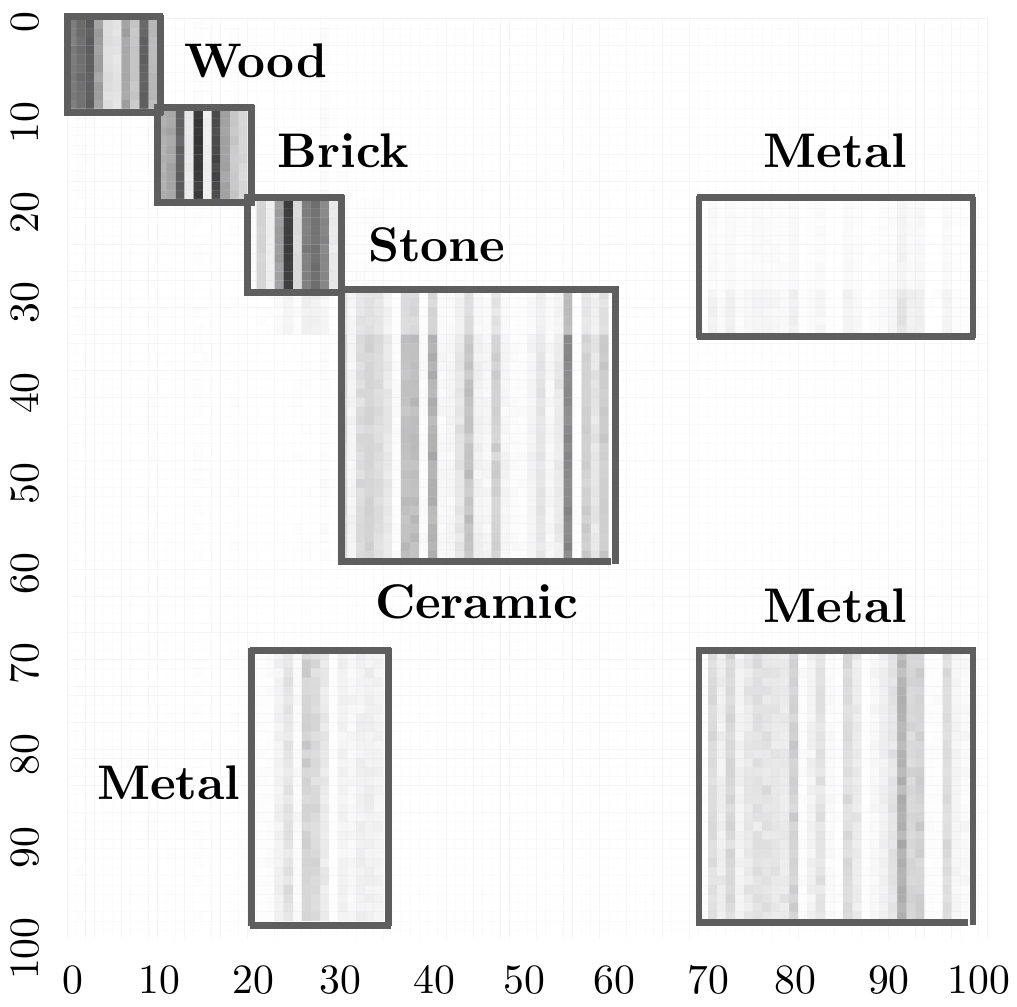}
    \caption{Texture only}
    \label{fig:blocks_confusion:a}
\end{subfigure}
\begin{subfigure}{0.32\linewidth}
    \centering
    \includegraphics[width=\linewidth]{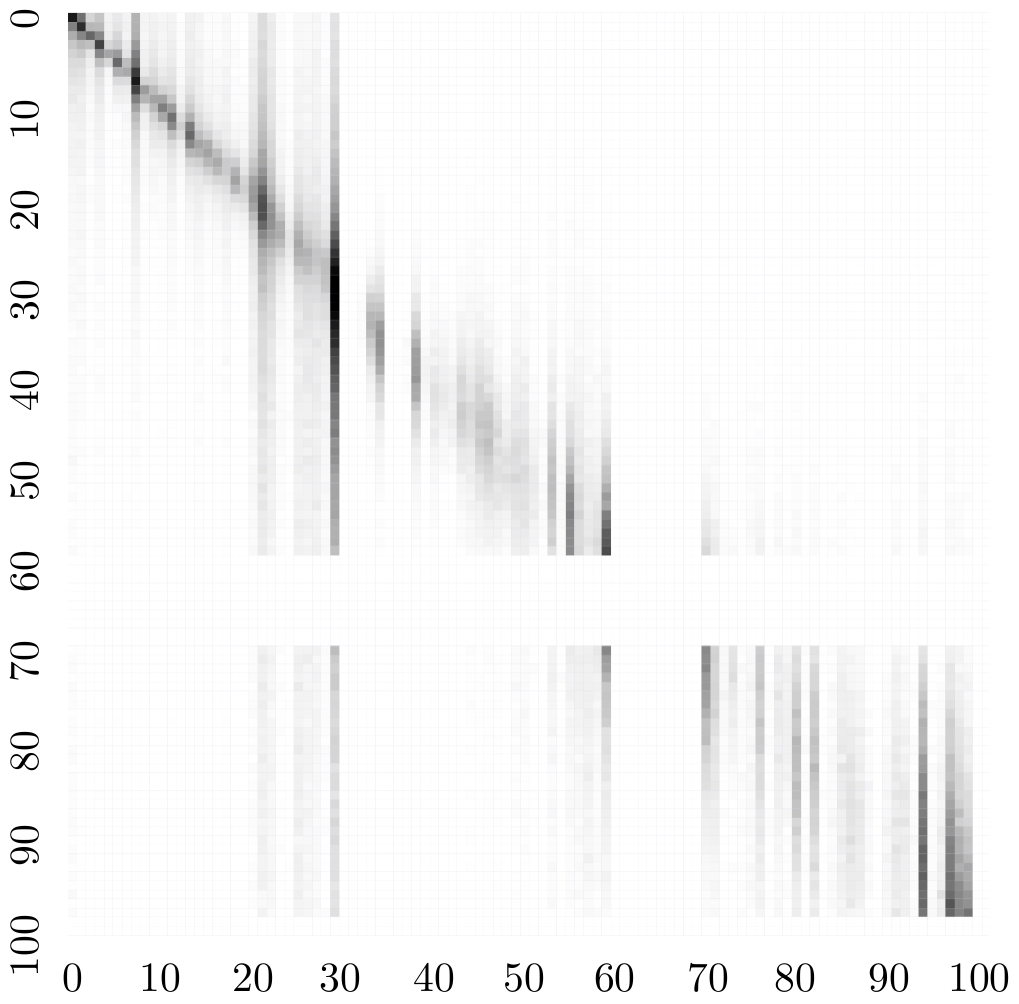}
    \caption{Physics only}
    \label{fig:blocks_confusion:b}
\end{subfigure}
\begin{subfigure}{0.32\linewidth}
    \centering
    \includegraphics[width=\linewidth]{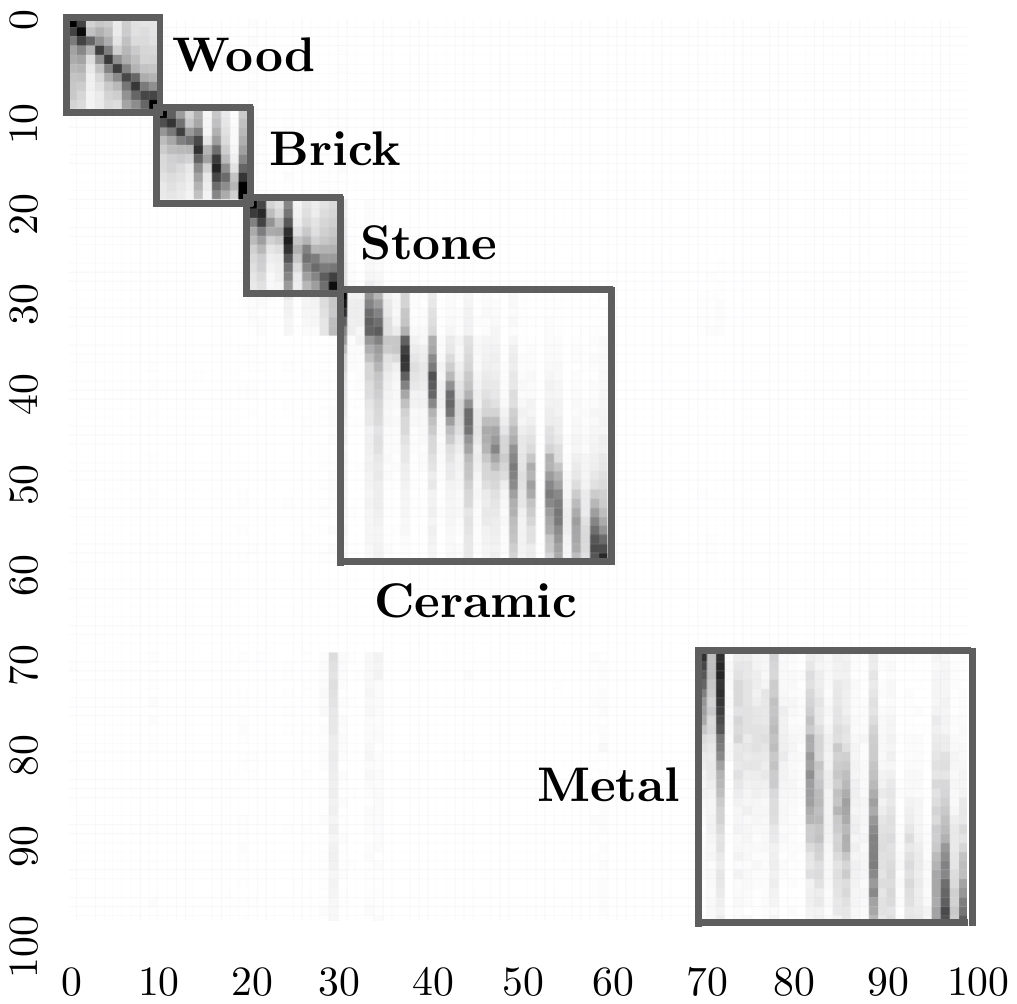}
    \caption{Texture + Physics}
    \label{fig:blocks_confusion:c}
\end{subfigure}

\vspace{1mm}
\includegraphics[width=0.95\linewidth]{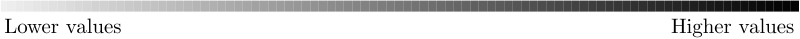}
\vspace{-8pt}
\caption{Confusion matrices of physical parameter estimation. The information provided by two types of observations are different: \textbf{(a)} with texture as input, our model tends to guess within the material's possible density values (see \tbl{tbl:material}); \textbf{(b)} with physics as input, our model only makes errors between close values.}
\label{fig:blocks_confusion}
\vspace{-20pt}
\end{figure}

\myparagraph{Preference over Possible Values}
We illustrate the confusion matrices of physical parameter estimation in \fig{fig:blocks_confusion}. Although our \model model performs similarly either with only texture as input or with physics as input, its preferences over all possible values turn out to be quite different. With texture as input (in \fig{fig:blocks_confusion:a}), it tends to guess within the possible values of the corresponding material (see \tbl{tbl:material}), while with physics as input (in \fig{fig:blocks_confusion:b}), it only makes errors between very close values. Therefore, the information provided by two types of inputs is orthogonal to each other (in \fig{fig:blocks_confusion:c}).

\myparagraph{Impact of Primitive Numbers}
As demonstrated in \tbl{tbl:blocks_comparison}, the number of blocks has nearly no influence on the model with texture as input. With physics interactions as input, the model performs much better on fewer blocks, and its performance degrades when the number of blocks starts increasing. The degradation is probably because the physical response of any rigid body is fully characterized by a few object properties (\ie, total mass, center of mass, and moment of inertia), which provides us with limited constraints on the density distribution of an object when the number of primitives is relatively large.

\begin{figure}[t]
\centering
\includegraphics[width=\linewidth]{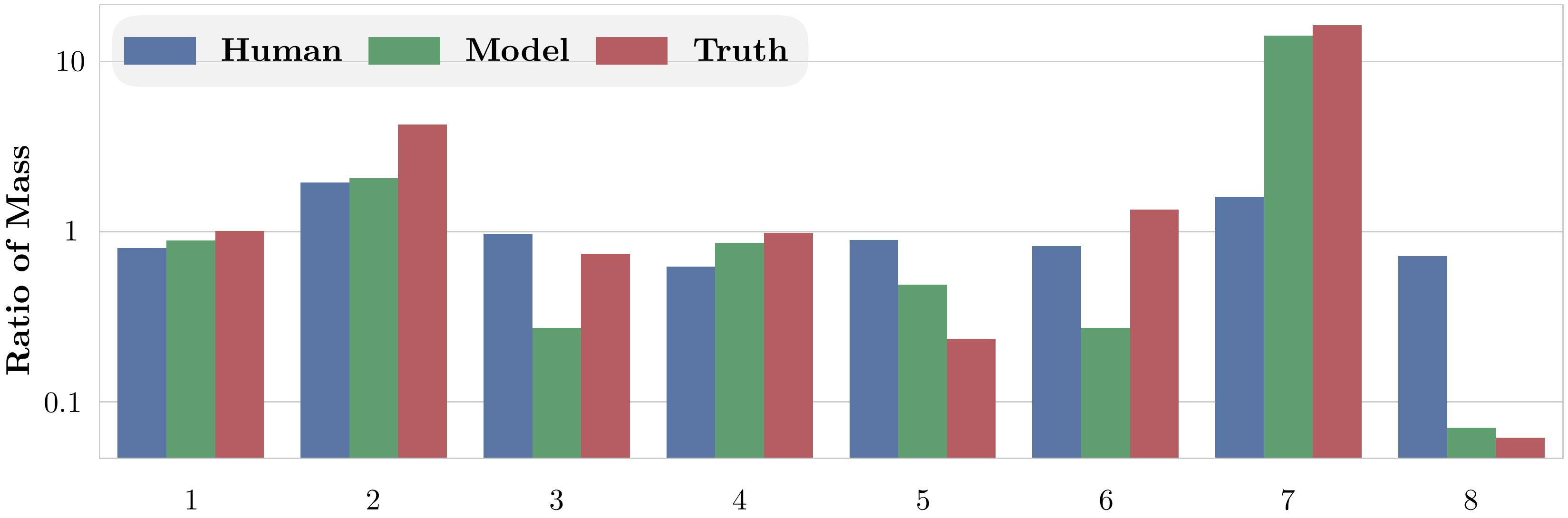}
\vspace{-18pt}
\caption{Human's, model's and ground-truth predictions on ``which block is heavier". Our model performs comparable to humans, and its response is correlated with humans.}
\label{fig:humans_phys}
\vspace{-8pt}
\end{figure}

\begin{table}[t]
\setlength{\tabcolsep}{3pt}
\centering\small
\begin{tabular}{cC{1.5cm}C{1.5cm}C{1.5cm}C{1.5cm}C{1.5cm}}
    \toprule
    Observation & 2 blocks & 3 blocks & 4 blocks & 5 blocks & Overall\\
    \midrule
    Texture & 18.2 & 18.5 & 18.8 & 19.7 & 19.1 \\
    Physics & 3.6 & 7.9 & 15.8 & 20.0 & 14.7 \\
    \cmidrule{1-6}
    Texture + Physics & 2.3 & 4.9 & 7.8 & 10.9 & 8.0 \\
    \bottomrule
\end{tabular}
\vspace{3pt}
\caption{Quantitative results (RMSE's) on block towers (with different block numbers): \textbf{(a)} with texture as input, our model performs comparably on different block numbers; \textbf{(b)} with physics as input, our model performs much better on fewer blocks.}
\label{tbl:blocks_comparison}
\vspace{-28pt}
\end{table}

\myparagraph{Human Studies}
We select the block towers with two blocks from our real dataset, and study the problem of ``which block is heavier" upon them. The human studies are conducted on the Amazon Mechanical Turk. For each block tower, we provide 25 annotators with an image and a video of physics interaction, and ask them to estimate the ratio of mass between the upper and the lower block. Instead of directly predicting a real value, we require the annotators to make a choice on a log scale, \ie, from $\{2^k \mid k = 0, \pm1, \ldots, \pm4\}$. Results of average human's predictions, model's predictions and the truths are shown in \fig{fig:humans_phys}. Our model performs comparably to humans, and its response is also highly correlated with humans: the Pearson's coefficient of ``Human \vs Model", ``Human \vs Truth" and ``Model \vs Truth" is 0.69, 0.71 and 0.90, respectively.

\section{Conclusion}

In this paper, we have formulated and studied the problem of physical primitive decomposition (\model), which is to approximate an object with a set of primitives, explaining its geometry and physics. To this end, we proposed a novel formulation that takes both visual and physics observations as input. We evaluated our model on several different setups: synthetic block towers, synthetic tools and real-world objects. Our model achieved good performance on both synthetic and real data.

\vspace{5pt}
\myackparagraph{Acknowledgements}
This work is supported by NSF \#1231216, ONR MURI N00014-16-1-2007, Toyota Research Institute, and Facebook.

\bibliographystyle{splncs04}
\bibliography{partphys}


\appendix
\renewcommand{\thesection}{A.\arabic{section}}
\renewcommand{\thefigure}{A\arabic{figure}}
\setcounter{section}{0}
\setcounter{figure}{0}

\section{Implementation Details}
\label{sec:model:detail}

We present some implementation details about network architecture and training.

\myparagraph{3D ConvNet}
As the building block of voxel encoder, this network consists of five volumetric convolutional layers, with numbers of channels $\{1, 2, 4, 8, 16\}$, kernel sizes 3$\times$3$\times$3, and padding sizes 1. Between convolutional layers, we add batch normalization~\cite{Ioffe2015Batch}, Leaky ReLU~\cite{Maas2013Rectifier} with slope 0.2 and max-pooling of size 2$\times$2$\times$2. At the end of the network, we append two additional 1$\times$1$\times$1 volumetric convolutional layers.

\myparagraph{Network Details}
As the inputs fed into different encoders, voxels $V$, images $I$ and trajectories $T_k$ are of size 1$\times$32$\times$32$\times$32, 3$\times$224$\times$224 and 256$\times$7, respectively. The dimensions of output features from encoders, $f_\text{V}$, $f_\text{I}$ and $f_\text{T}$, are all 64. Inside both trajectory encoder and primitive generator, we employ the Long Short-Term Memory (LSTM) cell with hidden sizes of 64 and dropout rates of 0.5 as recurrent unit. The trajectory encoder uses a single-layer recurrent neural network, while the primitive generator applies three layers of recurrently connected units.

\myparagraph{Training Details}
We implement our \model model in PyTorch\footnote{\url{http://pytorch.org}}. For the image encoder, we make use of the weights of ResNet-18~\cite{He2015Deep} pre-trained on ImageNet~\cite{Deng2009Imagenet:}  and replace its final classification layer with a fully-connected layer, while for other modules, we initialize their weights randomly. During optimization, we first train the geometric parameters (by setting $\omega$ to 0), and then we train all parameters jointly. Optimization is carried out using ADAM~\cite{Kingma2015Adam:} with $\beta_1 = 0.9$ and $\beta_2 = 0.999$. We use a learning rate of $\gamma = 10^{-3}$ and mini-batch size of 8.

\end{document}